# A deep learning approach to early identification of suggested sexual harassment from videos

Shreya Shetye[1,2], Anwita Maiti[1], Tannistha Maiti[1] and Tarry Singh[1]

[1]Deepkapha AI, [2]Vellore Institute of Technology, Bhopal, India

Sexual harassment, sexual abuse, and sexual violence are prevalent problems in this day and age. Women's safety is an important issue that needs to be highlighted and addressed. Given this issue, we have studied each of these concerns and the factors that affect it based on images generated from movies. We have classified the three terms (harassment, abuse, and violence) based on the visual attributes present in images depicting these situations. We identified that factors such as facial expression of the victim and perpetrator and unwanted touching had a direct link to identifying the scenes containing sexual harassment, abuse and violence. We also studied and outlined how state-of-the-art explicit content detectors such as Google Cloud Vision API and Clarifai API fail to identify and categorise these images. Based on these definitions and characteristics, we have developed a first-of-its-kind dataset from various Indian movie scenes. These scenes are classified as sexual harassment, sexual abuse, or sexual violence and exported in the PASCAL VOC 1.1 format. Our dataset is annotated on the identified relevant features and can be used to develop and train a deep learning computer vision model to identify these issues. The dataset is publicly available for research and development.

## 1. Introduction

Sexual violence is a serious and widespread issue that affects individuals of all ages, genders, and backgrounds. According to the World Health Organization (WHO), approximately 1 in 3 women worldwide have experienced physical or sexual violence. In addition, men and boys can also experience sexual violence, although it is less commonly reported. LGBTQ+ individuals are more likely to experience higher violence.

A study conducted in the United States found that about 20% of women and 5% of men reported experiencing sexual assault at some point in their lives. Another study in Canada found that approximately 1 in 3 women and 1 in 6 men had experienced sexual violence in their lifetime.

In the past few years, video analytics, also known as video content analysis or intelligent video analytics, has attracted increasing interest from both the industry and the academic world. Thanks to the popularization of deep learning, video analytics has introduced the automation of tasks that were once the exclusive purview of humans.

Recently video analytics has been providing a wide range of applications in monitoring traffic jams and providing alerts in real-time, analyzing customers' flow in retail to maximize sales [3]. Its been also used in more well-known scenarios such as facial recognition or smart parking. It can be used similarly to detect and prevent violence in a variety of settings, including schools, public transportation systems, and public spaces. Identifying sexual harassment such as stalking which is following or laying in wait at places such as home, school, work, or recreational spaces through CCTV footage, video analytics is a study that can have a wide range of benefits. In real-time images, computer vision can identify unwanted physical touching or advances, persistent attempts to engage in conversation or make gestures, and invade private space, and facial expressions of the victim or perpetrator. Even sounds like hooting or suggestive comments

*Corresponding author(s): tannistha.maiti@deepkapha.com*

can be identified through advanced deep-learning techniques.

We believe it is timely and important to systematically investigate sexual violence in images and understand its factors, based on which automatic detection approaches can be formulated. We discuss the prevalence and impact of sexual harassment, abuse, and violence on women. In this work, we first collect a dataset of sexual violence images from six Indian movies which are labeled by a social scientist. We analyze the images in our dataset against two state-of-the-art explicit image detectors, Google Cloud Vision API, and Clarifai NSFW. Then, we study the sexual violence images in our dataset to determine the visual factors that are associated with such images. The dataset provides a unique opportunity to study the representation and portrayal of these issues in popular media and to analyze the social and cultural contexts in which they occur.

The key contributions of this paper are as follows:

- New Dataset of Sexual Violence Images. We present the methodology to collect a large dataset of 500 images of sexual violence images was created based on frames of six movies. The dataset with images has been annotated by a social scientist for attributes such as facial expressions and unwanted touching. The images are then tagged as sexual abuse, sexual harassment, and sexual violence.
- We present a measurement of two state-of-the-art offensive image detectors against our images dataset, wherein we study their effectiveness in detecting the content of images. We find that these state-of-the-art detectors are not capable of effectively identifying sexual violence in images.
- We present the interview conversations gathered from students in various US universities and their views on this system also in aspects of ethical concerns.

Our dataset is publicly made available and anyone can download the dataset with annotations at this link

## 2. Terminologies

The multitude of experiences encompassing sexual violence makes it a challenging and complex phenomenon to explore.

In the context of sexual violence, labeling the situation is complicated by the variety of terms and descriptors people can use to refer to their lived experiences [1]. Yet, Researchers, counselors, and the media tend to use different terms to describe sexual violence committed against women. For example, a rape crisis center had an unquestioning acceptance of some words (i.e., sexual assault and survivor) in lieu of others (i.e., rape and victim), which implies that there are "right" and "wrong" ways to talk about sexual violence [9]. Specifically, sexual events encompass incest, molestation, and rape. Rape is defined as any form of sexual contact committed against an individual's will.

### 2.1. Sexual Assault

Sexual assault occurs when you are the victim of intentional physical contact that is sexual in nature without your consent. This can include unwanted sexual touching, rape, and other similar acts. Sexual assault has been defined as being pressured or forced to have sexual contact [4]. Rape and sexual assault have been used interchangeably in coverage of events leading to the **#MeToo** movement, and this practice, though unintentional, is confusing. However, in this study, we do not consider the actions that lead to rape as an assault but rather as sexual violence [8, 2].

### 2.2. Sexual Harassment

Sexual harassment includes unwanted sexual conduct such as unwelcome sexual advances, verbal conduct, physical conduct, and requests for sexual favors in instances. The conduct unreasonably interferes with the ability to do a job or creates an intimidating or hostile work environment. This situation might include the displays of offensive material or inappropriate jokes on an ongoing basis.

## 2.3. Sexual Violence

Social and behavioral scientists often use the term "sexual violence." This term is far broader than sexual assault. It includes acts that are not codified in law as criminal but are harmful and traumatic. Sexual violence includes using false promises, insistent pressure, abusive comments, or reputational threats to coerce sex acts. In this study, we consider sexual violence as an aggressive form of the sexual act often similar to actions of rape.

## 2.4. Victim and Perpetrator Survivor

The two most commonly used labels are victim and survivor. A victim typically is believed to be someone injured, killed, or otherwise harmed by some individual, act, or condition. The term victim captures the sense of injury and injustice felt by individuals who have experienced a sexually [7]. The term perpetrator refers to any person who commits a sexual assault, regardless of whether the victim is a minor or an adult [5].

## 3. Violence detection API

### 3.1. *Google API*

Explicit Content Detection in the Google Cloud Video Intelligence API, which detects adult content in images and videos, was used. It categorizes the pornography likelihood of the clip among very unlikely, unlikely, possible, likely, and very likely. When Explicit Content Detection evaluates a video, it does so on a per-frame basis and considers visual content only. The audio component of the video is not used to evaluate explicit content level. The Vision API SafeSearch detection feature uses a deep neural network model specifically trained to classify inappropriate content in images. It computes a score on a 0 to 1 scale and, based on that score, applies the corresponding likelihood as a string. The result is in the form of a bucketized likelihood value from very unlikely to very likely, depending upon the level of explicit content detected in the particular frame along with the time offset of the frame. The frames categorized as possible, likely, and very likely were taken into interest for dataset creation.

### 3.2. *Clarifai*

In Clarifai, we used the moderation recognition API, which rates the probability of safe, suggestive, drug, explicit, and gore in each frame. The Clarifai API returns a response that lists the predicted concepts for every video frame. The Image Moderation Classifier model is used on every frame to moderate nudity, sexually explicit, harmful, or abusive user-generated imagery (UGC). The Clarifai Image Moderation Classifier model utilizes an Inception V2 architecture and is trained using their proprietary dataset. The model outputs a probability distribution among five different labels, namely, drug, explicit, gore, safe, and suggestive. Their website and documentation fail to provide a clear definition of each of the categories. Frames with a high probability of suggestive, explicit, and gore were taken into account for the dataset creation.

We observed that these state-of-the-art detectors fail to detect sexual harassment on multiple instances. The poor performance of these detectors can be attributed to the fact that they have not advertised to detect sexual harassment or sexual violence.

We can first observe that none of the state-of-the-art detectors consider sexual harassment or sexual violence in images or videos as a category of offensive content. Thus, our first motivation is that this important category of offensive content should be included by existing systems as an offensive content category. Secondly, since the factors of sexual harassment in images are not very well defined, the existing detectors are not capable of detecting them. As a result, we're encouraged to elucidate on recognizing the visual characteristics of sexual harassment so that they may be identified automatically in photos.

We created graphs to better analyse the results of the Google Cloud API and the Clarifai API on the clips of the movies. Figure 1 illustrates a histogram based on the results of Explicit Content Detection by the Google Cloud Vision API on a clip from the movie Bandit Queen. The colour of the bar categorises the frame of the video ac-cording to the categories depicted in the upper right corner of the graph. These categories depict the likelihood of pornographic content in the particular frame of the video clip.

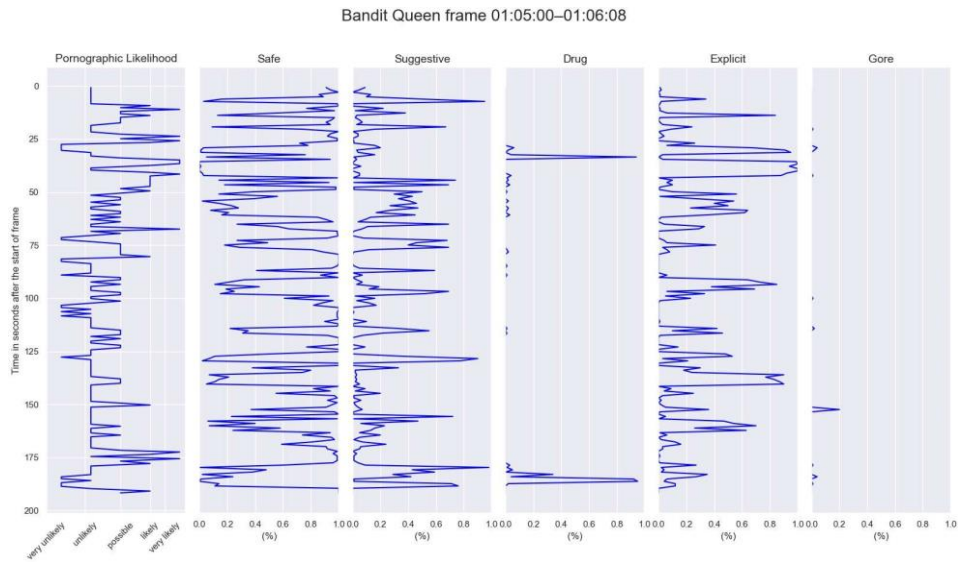

Figure 1 | The results of Google Cloud Explicit Content Detection API and Clarifai moderation recognition API on a clip from the movie Bandit Queen in the time frame 01:05:00-01:06:08. The factors used to analyze the frames are pornographic likelihood, safe, suggestive, drug, explicit, and gore. Google Cloud does a fairly decent job of identifying pornographic content. However, it does miss identifying certain scenes where blatant sexual violence is shown. Clarifai fails to recognize the gore in most of the scenes where there is a clear visual of blood on the victim. There is a false detection of a drug. It does miss to confidently identify several scenes containing sexual violence as explicit scenes and also identifies some as suggestive.

## 4. Sexual Violence Images Data

To identify factors of sexual violence in images, we collect data from Indian movies [6], which should be representative of real-world sexual violence. In our work, we first extract the frames of different movies identified to contain the graphic representations of sexual violence based on [6] Our data collection tasks are approved by a social scientist. These movies have depicted violence against women and are perfect candidates to identify the attributes that are associated with sexual violence. Table 1 provides a detailed explanation of the attributes that are evident from the movie clips. We further tag the images as sexual violence, sexual abuse, and sexual harassment. The scenes taken into account depict the acts of sexual harassment and sexual violence against women.

The dataset contains 210 images all annotated for 3 main features.

### 4.1. Annotation

In order to make the dataset appropriate to train and evaluate machine learning models, we carried out an annotation process on all 210 images. The annotation for this dataset was done using the CVAT annotation tool. Our objective was to mainly identify and annotate the victim, and the perpetrator as well as any visible signs of sexual harassment. The annotations on the images of the dataset include the victim, perpetrator, and unwanted touching. For annotating the victim and the perpetrator, the faces and expressions of the people were taken into consideration. Facial expressions such as excitement and anger were used to identify the perpetrator. To identify the victim, facial expressions such as fear and anxiety were highlighted. The facial expressions of the people in context were segmented by drawing a bounding box enclosing their entire heads. For annotating unwanted touching, the unwelcome touches of the perpetrator on the victim were marked using bounding boxes. These bounding boxes were then labeled with the corresponding class labels. Multiple annotations were done on each image of the dataset. Each image was also tagged in one of the 3 categories, namely, sexual harassment, sexual abuse, and sexual violence. With the help of our definitions provided for each of these terms, each image was tagged accordingly. The annotated dataset was then exported in the PASCAL VOC 1.1 format.

| Movie | Year | Director | Timestamp | Nature |
|---|---|---|---|---|
| Dastak | 1970 | Rajinder Singh Bedi | 01:01:56–01:02:49 | Marital Rape |
| Adalat O Ekti Meye | 1982 | Tapan Sinha | 00:28:30–00:32:30 | Gang Rape |
| Damini | 1993 | Rajkumar Santoshi | 00:42:44–00:44:58 | Gang Rape |
| Bandit Queen | 1996 | Shekhar Kapur | 01:05:00–01:06:08 | Marital Rape and Gang-rape |
| Bulbbul | 2020 | Anvita Dutt Guptan | | |
| Matrubhoomi | 2003 | Manish Jha | 00.51.01–00.52.05 | Marital Rape and Gang Rape |
| Matrubhoomi | 2003 | Manish Jha | 01.10.35–01.11.36 | Marital Rape and Gang Rape |

Table 1 | List of all the movies used in creating the data. The timestamps used in the movies are also mentioned.

### 4.2. Pre-processing

Certain images in the dataset were too dark to decipher any information from them. In order to solve this problem, we applied some pre-processing techniques so that they can be perceived with ease. This will also help to aid the process of detecting and identifying the features through the machine learning model. For this, we first applied image equalization (contrast enhancing) to the graph. In order to enhance the contrast to a certain extent, we used the technique called histogram equalization. Histogram Equalization is a contrast-enhancing computer image processing method. It does so by effectively spreading out the most common intensity values, i.e., expanding out the image's intensity range. When the useful data is represented by near-contrast values, this approach frequently boosts the global contrast of pictures. This enables locations with poor local contrast to obtain an increase in contrast. The resulting images were significantly brighter and more decipherable than the original images.

## 5. Use Case importance

We interviewed US university students who were either victim of sexual assault or concerned about the problem. Based on the discussion on the usefulness of identifying situations such as sexual abuse, sexual harassment, and sexual violence, we gathered significant concerns. Students, in general, like the idea of deterrents. They felt that real-time deep-learning computer vision could provide solutions in situations like stalking. Students would feel safer in locations like school libraries if an app that identifies sexual harassment situations in real time were present. If situations escalate, immediate involvement of security can be provided. According to [6], many movies either depict or "celebrate" rape in a detailed manner, and it is disturbing to note that in the name of trying to contribute to women's emancipation, instead end up doing the opposite by glorifying sexual violence on screen, making the situation worse for women. A deep learning model trained on this dataset should be able to identify whenever a movie depicts these scenes and flag them as inappropriate. This dataset can be used to build a detector that can identify scenes containing sexual harassment, sexual abuse, and sexual violence in videos. Such a detector can be used to monitor videos posted online on family-friendly websites. Any video uploaded on the website should first pass through the detector to ensure that the content does not contain any explicit or suggestive scenes. Warnings can be issued based on the confidence level of the detector. This will ensure that the content uploaded on the website is safe. Additionally, it can be incorporated as a web extension that can detect such scenes in videos. Upon detecting the scenes, the extension can provide a trigger warning and an option to skip the scenes containing disturbing content in the video.

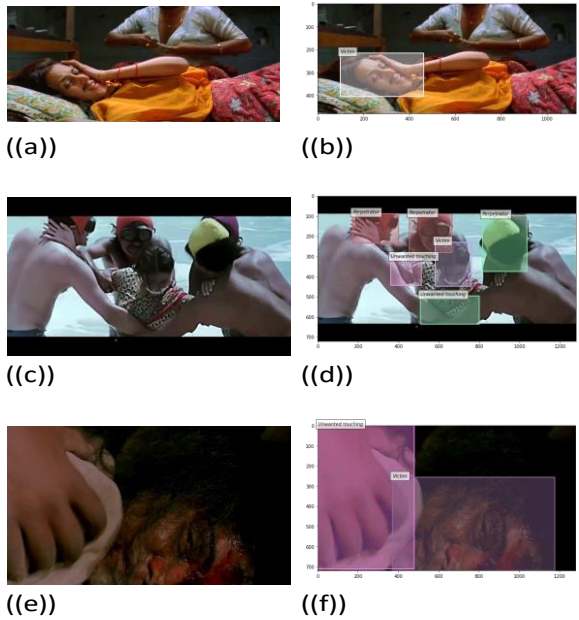

Figure 2 | (a): Image from the dataset depicting sexual harassment. (b): Image (a) annotated to show the victim. (c): Image from the dataset depicting sexual abuse. (d): Image (c) annotated to show the victim, perpetrators, and unwanted touching. (e): Image from the dataset depicting sexual violence. (f): Image (e) annotated to show the victim and unwanted touching.

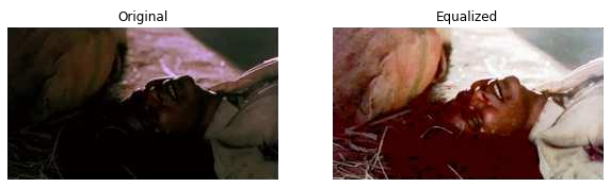

Figure 3 | Histogram equalization is applied to the original image to reduce the contrast and make the image more discernible.

## Conclusion

In this paper, we study the phenomenon of sexual abuse, harassment, and violence through images, specifically factors such as facial expressions and physical interaction between the victim and perpetrator. We have discussed how images have content of sexual violence and have identified two major factors. We also provided our description of sexual violence, abuse, and harassment. For the first time, we present annotated and tagged dataset that can eventually be used in deep learning for sexual violence identification. To emphasize the importance of deep learning identification in sexual harassment, we have also interviewed university students and identified the applicability of the deep learning model that can be developed with this dataset. As part of future studies, we propose implementing a deep-learning classifier to detect facial expressions and unwanted touching. We will also further look into the effectiveness of the identified visual factors in terms of the models.

## Disclaimer

These images from the dataset contain violent content that may be disturbing to some readers. Reader discretion is advised. The scenes and actions depicted in this program are fictional and should not be imitated. Moreover, the portrayal of any character does not reflect the real-life beliefs or actions of the actor or any individuals associated with the production.

# 6. Appendix

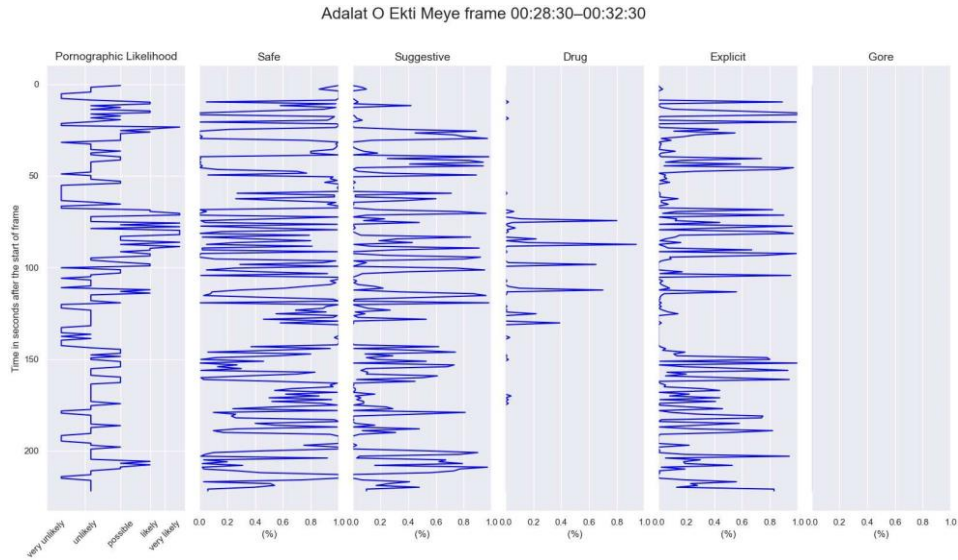

figure 4 | The results of Google Cloud Explicit Content Detection API and Clarifai moderation recognition API on a clip from the movie Adalat O Ekti Meye in the time frame 00:28:30-00:32:30. Both Google and Clarifai do not do a good job of identifying sexual abuse in the clip. Either detector does not flag several scenes of forcing and manhandling the victim. Scenes including sexual abuse are also not detected. Clarifai also has false detection of 'drug' in the clip.

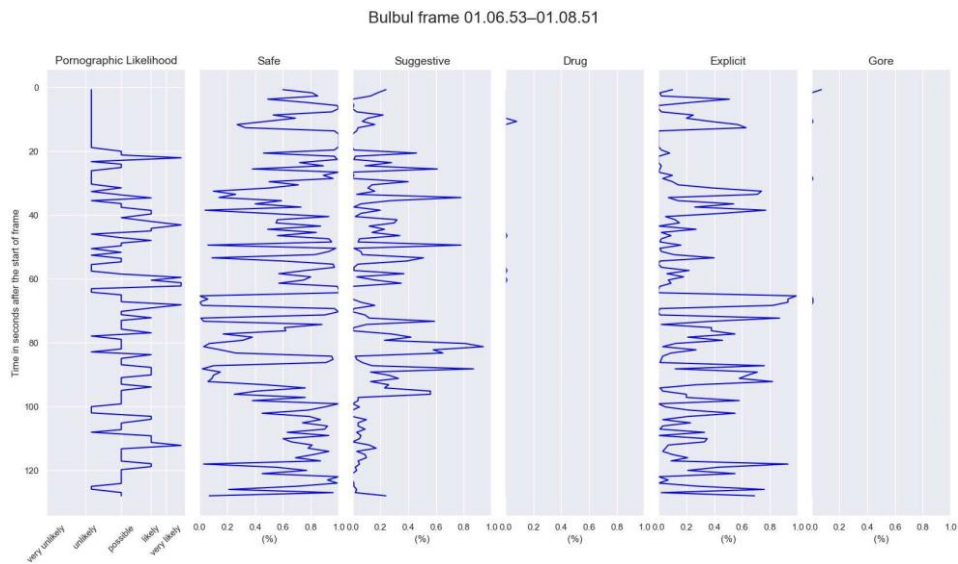

Figure 5 | The results of Google Cloud Explicit Content Detection API and Clarifai moderation recognition API on a clip from the movie Bulbul in the time frame 01:06:53-01:08:51. In this clip, most of the scenes of sexual abuse and violence are identified by both Google and CLarifai detectors. Google performs decently in identifying the pornographic content in the clip. Clarifai has very few false detections of 'drug' and 'gore'. Clarifai, however, has a high probability of 'safe' and a low probability of 'suggestive' and 'explicit' in most scenes containing sexual violence.

A deep learning approach to early identification of suggested sexual harassment from videos| Factors | Attributes | Sexual Abuse | Sexual Violence | Sexual Harassment |
|---|---|---|---|---|
| Body Posture of Perpetrator | Unwanted touching, hugging, kissing | Y | Y | |
| | Invading personal space and privacy through actions such as cornering and peeping | | | Y |
| | Invading personal space and privacy through actions such as breathing down one's neck | Y | | |
| Offensive Gesture of the Perpetrator | Deliberate brushing | Y | | |
| | Stroking own private parts | | | Y |
| | Stroking the private parts of a victim | Y | Y | |
| | Smacking lips | | | Y |
| | Elevator eyes, Ogling, Staring | Y | | Y |
| | Leering, Whistling, winking | | | Y |
| | Rape | | Y | |
| | Gore | | Y | |
| | Pinching, tugging at clothes, Tearing clothes | | Y | |
| | Manhandling | Y | Y | |
| | Threatening, Taunting | Y | | Y |
| | Physical abuse | | Y | |
| | Undressing self | Y | | Y |
| | Undressing Victim | Y | Y | |
| Facial Expression of Perpetrator | Excitement, Angry, Triumphant | Y | Y | Y |
| Facial Expression of Victim | Fear, Anxiety, Pain, Crying, Agonized | Y | Y | Y |

Table 2 | This table is based on our definitions. An important point to note here would be that these attributes are categorized according to when they are considered independently. When one or more different attributes are considered together, they can be categorized differently.

9